# A SYSTEM OF SERIAL COMPUTATION FOR CLASSIFIED RULES PREDICTION IN NON-REGULAR ONTOLOGY TREES

Kennedy E. Ehimwenma, Paul Crowther and Martin Beer

Communication & Computing Research Centre
Department of Computing
Sheffield Hallam University, United Kingdom

## ABSTRACT

*Objects or structures that are regular take uniform dimensions. Based on the concepts of regular models, our previous research work has developed a system of a regular ontology that models learning structures in a multiagent system for uniform pre-assessments in a learning environment. This regular ontology has led to the modelling of a classified rules learning algorithm that predicts the actual number of rules needed for inductive learning processes and decision making in a multiagent system. But not all processes or models are regular. Thus this paper presents a system of polynomial equation that can estimate and predict the required number of rules of a non-regular ontology model given some defined parameters.*

## 1. INTRODUCTION

Probabilistic models such as in Bayesian nets and Markov network, models decision processes based on some measurable parameters that are for instance discrete (true or false) from state-to-state or node-to-node transition. Typically, a network is a graphical representation with a collection of nodes representing random variables and edges connecting the nodes [28]. While Bayes nets are of directed graphs, Markov are undirected.

Objects or network structures that are regular take uniform dimensions. But not all processes or models are regular.While the work of [16], [17] projected the *Initialisation* algorithm for regular ontology number of rules prediction, this paper projects a polynomial for the non-regular ontology counterpart. The polynomial equation predicts the number of classified induction rules in a multiagent system(MAS) for the classification of students for appropriate learning material(s) after pre-assessment about their SQL learning. As a MAS system developed on the platform of Jason AgentSpeak language [6]: description logic based language, the classification—*triggering_e*vent-*condition-action*—rules are represented in a classifier agent in predicate logic formalism.Pre-assessment is performed based on the boolean condition probabilities parameters, and thereafter inter-agent communication for learning material recommendation based on the given ontology concepts and their relational properties. Generating a number of probability distribution for condition-action learning is a huge task at the knowledge acquisition stage for humans especially for some big or exponentially increasing ontology trees. As a result, different algorithms or methods have been proposed for conditional probability learning (CPT) for uncertainty reasoning in Bayesian or semantic nets [12], [13], [33], [11].

Bayesian network are acyclic directed graphs whose nodes are random variables $X_1, ..., X_n$, and a collection of conditional probability distributions, one distribution for each random variable given its parents[27]. Bayes net consists of two parts: a directed acyclic graph that represents direct influences among variables, and a set of *conditional probability tables* (CPT) that quantify the strengths of these influences [12]. Thus Probabilistic Graphical Models are powerful and effective tools for representing and reasoning under uncertainty [21], [8]. Ontology based decision methods leverage greater understanding of semantic relationships by applying probability logic e.g. fuzzy type 1 and fuzzy type 2to tackle vagueness[4], [29]. This emphasises the important role of ontologies in the development of knowledge based system which describes semantic relationships among entities [30].

As stated earlier, in Bayes nets, probabilistic relationships otherwise known as properties are represented on the directed-edges to predict the casual relations between a current node and the next direct node. Now, within a given belief network or ontology structure, how can the total amount of classified induction rules for both positive (true) and negative (false) training examples be estimated and predicted? In [16], [17] directed edges represented *object properties* [19] of some semantic predicate and knowledge representation between nodes in the ontology. The edges as directed path for student learning through to the leaf-node terminals where students' learning are pre-assessed and classified based on two possible state (boolean) parameters. In this work, we consider only boolean variables: we ascribed to each variable *Xi* the values 1 ("pass") and 0 ("fail") states; in which only one of the states can be attained per node for execution.

The work of [16], [17] formulated some system of algorithmic equations for predicting the required number of classified induction rules for agent classification learning in regular ontologies. But in this paper, the focus is on the estimation and prediction of rules in non-regular ontology trees. The paper presents a system of polynomial equation that can predict the number of induction rules for non-regular ontologies. This supports the knowledge or system engineer to know beforehand the number of inductive learning clauses that is needed for a given system. The remaining part of this paper continues with related work in Section 2. Section3 describes the system of polynomial equation for induction rule estimation and prediction in non-regular ontologies using *parent-to-child*node relationship analysis. In Section4 knowledge representation of the estimated rules are presented. Section5 is conclusions and further work.

## 2. RELATED WORK

This Section describesome work on classification learning using boolean value parameters, and prediction models for probabilistic uncertainty learning with regards to semantic nets.

### 2.1 Reasoning in Semantic Nets and Recommender Systems

Various computational techniques exist for conceptual estimation of ontology tree structures. But learning about probabilistic uncertainties take different forms. While some are of the fuzzy sets distribution, others are discrete or continuous Gaussian distribution [11]. From the usage ofsemantic properties and node attributes, semantic based recommender systems have been exploited to the advantage of making or generating recommendation. This is because the underlying techniques provide a means of discovering and classifying unknown features or items [32],[31]. Based on semantic nets, some classification or recommender systems have been built on user-to-user relations [1]or/anditem-to-item relation [25], [14]. These are systems where recommendation process is determined by user rating of items—positive or negative—which are used to calculate the *similarity* between a given set of *n* features in a neighbourhood. An example is the conceptual similarity computational algorithm that can recommend tree structured items based on the users' fuzzy item preference tree and matching to item trees [32]. This type of uncertainty reasoning has also been applied to learning in ontology based decision support systems e.g. Ontology-supported case-based reasoning (OS-CBR) [2]. The classification learning estimation process in this paper is based on user-to-item i.e. students' desired_node-to-leaf_node relations.

## 2.2 Probability Inference Computation

Polynomial linear models have been proposed to estimate probability distribution over some semantic nets. This is when there are some prescribed variables and network nodes. Variables can be either continuous or discrete [11],[23]. Cutic [11] states that for *discrete* variables, each variable $X_i$ and its parent $Y_i$ can have their proportional probability distribution represented as a *conditional probability table* (CPT). That, if $n_1, …, n_k$ are the number of possible values of variables $Y_i$ respectively, the CPT for variable $X_i$ will have $\prod_{i=1}^{k} n_i$ entries. But as asserted by [1] there is a downside: that the number of free parameters is exponential in the number of parents.

In [12] a network polynomial linear model was also presented which was shown to compute conditional probabilities using *arithmetic circuits,* and retrieved answers to probabilistic queries by evaluation and partial derivative of the polynomial terms. The network polynomial which was multivariate in function was demonstrated with Boolean parameters *[true or false]* with assigned fuzzy set values between the range [0,1] for every parent-child relation. Darwiche [12] states that the polynomial was exponential in size and so could not be represented explicitly. Hence, its representation as a system of multlinear arithmetic circuits. The CPT [5], [12], [13], [15], [18], [24], [9], [3] are usually used to specify the set of probability distribution. In Das (2004) for instance, a computational model for CPT was presented and computed the values of a CPT using a linear *weighted* sum algorithm.The algorithm was demonstrated with 5 learning features (or variables) over a semantic net of 3 parent nodes with converging edges to 1 child node. The work of [22] used CPT over ontology based semantic network to represent decision-making probabilities for disease diagnosis. To make prediction, two-state boolean parameters *[true or false]* were used for parameter learning in the *main* Bayes net (BN) and 3 parameterised features for the *sub* BN.

Similarly, joint probability distribution algorithm have also been applied in decision support inferencing (e.g. [7], [9], [20], [10], but to the best of our knowledge there is none for the estimation and prediction of the actual number of required probabilistic induction rules or classification learning based on semantic ontology trees.We believe that when the required number of rules are known at the knowledge acquisition stage, then classification learning can be modelled easily without missing any classification category whether manually or automatically. This paper thus presents a system of polynomial equation for estimating and predicting the number of probabilistic inference or induction rules. The probabilistic distribution are of an exponentially scalable polynomial equation which is a classification learning model, tested and proven based on two possible (or boolean) state classification parameter *T = 2.*This allows the predictive learning of uncertainties of probabilistic distribution over a semantic net.

## 2.3 Discrete Parameters and Classification

Discrete or boolean parameters are values that takes only two possible values: 1 or 0. In practical systems this implies: *on* or *off*, *good* or *bad, working* or *notWorking, found* or *notFound, present* or *notPresent, etc.* The investigation of relational-concept ontology in [16], [17] for the pre-assessment of students from node-to-node in a *regular ontology:* ontology with equal number of leaf-nodes across all parent nodes (Fig.1) used the boolean parameter *T = 2* torepresent either a *pass* or a *fail for* measurement of pre-assessment outcomes. The parameters *T* and leaf-nodes *N* of a given parent class node then formed the basis of agent classification learning. Classification learning is a *feature*, *instance* or *attribute* learning. Classification problem as stated in [26] consists of taking input vectors and deciding which *N* classes they belong to, based on training from exemplars of each class—where each example belongs to precisely one class.

### 2.4 The Initialisation Algorithm

To estimate and predict the number of classified induction learning rules R, [16], [17] as stated earlier projected the ***Initialisation*** algorithm

$$R = CT^N + 1 \qquad eq.1$$

for regular ontologies, on a multiagent based Pre-assessment System. This paper thus takes this work further into the concept of a *non-regular* ontology and classified number of induction rules prediction.

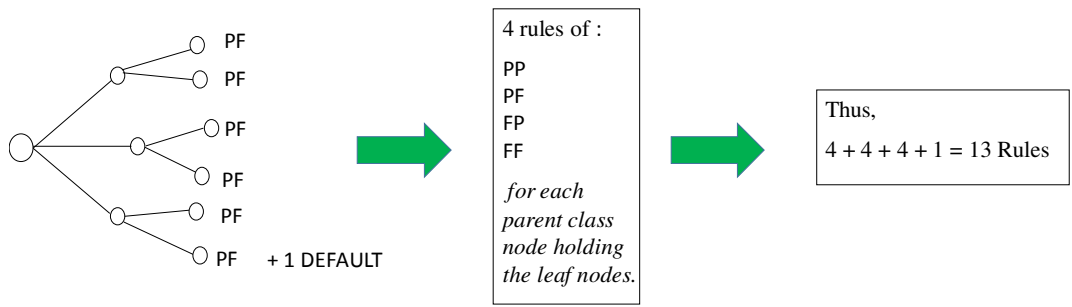

Fig. 1: A regular ontology

In Figure 1, the *PP*, *PF*, *FP*, *FF*(the set of 4) each are the probability distribution for each parent and their 2 leaf nodes relation,plus 1 *default rule* (for the last parent node with no prerequisite or underneath node) represent the set of selective classification pair. The P ≡ 1 and F ≡ 0.

## 3. RULES ESTIMATION AND PREDICTION IN NON- REGULAR ONTOLOGY TREES

In contrast to the concept of regular ontologies, a non-regular ontology tree structure is an ontology with a random structure and arbitrary number of leaf nodes. Given the exemplar Figure 2(a) & (b), how can classified inductive learning rules be estimated and predicted for *non-regular ontologies?* One of the probable solution is to take apart every *relationship* of *parent-class* and their *leaf-nodes* and treat them as unit components of the sum.To estimate the required number of rules R, we need to extend the ***Initialisation*** equation and apply it as a system of *polynomials* with each term in the equation representing an individual unit-node and their relation

*relationship(parent_class, leaf-node)*

in a 1:N relationship. That is, a parent-class to its respective leaf-node terminal(s). Therefore, from our scalable *Initialisation* equation *eq.1*, we rewrite

International Journal of Artificial Intelligence & Applications (IJAIA), Vol. 7, No. 2, March 2016

$$R = 1 + CT^N \qquad eq.2$$

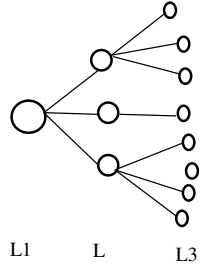
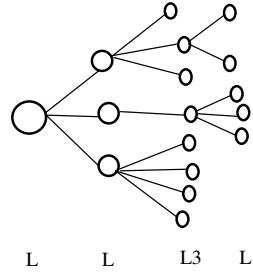

Fig. 2(a): A Non-Regular 3 Levels     Fig. 2(b): A Non-Regular 4 Levels

For the *non-regular* ontology tree with hierarchies of a number of prerequisite parent class $C$ (such as in Figure 2) and leaf-node $N$:

$$N \in \{1, 2, 3, \ldots, k\},$$

the number of classified rules $R$ can be estimated as

$$R = 1 + C_1 T^{N_1} + C_2 T^{N_2} + C_3 T^{N_3} + \ldots + C_k T^{N_k} \qquad eq.3$$

where the *constant 1* in eq.3 is the default--ground---rule for the lowest class concept that has no prerequisite class node, and the subscripts of $C$ in the terms represents position values of each component parent class node or unit of the tree. In summation, eq.3 becomes

$$R = 1 + \sum_{x=1}^{k} C_x T^{N}_x \qquad eq.4$$

where

$$x = N = \{1, 2, 3, \ldots, k\}$$

and $C = 1$ irrespective of the position of the node; and the subscript $x$ is unordered.

As shown, every parent class $C$ in eq.3 is addressed as a standalone node relative to some leaf-node(s). This enables the computation of the total amount of classified rules $R$ in a given ontology tree irrespective of the shape or size of the ontology.

### 3.1 Computational Illustrations and Prediction

Comparatively, Figure 1 depicts the structure of a regular ontology and Figure 2 (a) & b) those of non-regular ontologies. In Figure 2(a), there are three Levels of hierarchy L1 (root node), L2 (first parent class node), and L3 (terminal leaf-nodes). At Level L2 there are three parent class node with random number of leaf-nodes. In Figure 2(b) are four Levels: The root node level (L1), the leaf-nodes level (L4) and parent/leaf-node levels L2 and L3 in between.

Given that the parameter $T = 2$: a two boolean possible states *[0 or 1] for fail and pass respectively in a teaching-learning paradigm*; the total number of classified induction or decision rules $R$ for a non-regular ontology (Fig. 2(a)) can be computed as:

25

**Example 1:**

Thus,
$$R = 1 + \sum_{x=1}^{k} C_x T^N_x$$

$$R = 1 + C_1 T^N_1 + C_2 T^N_2 + C_3 T^N_3$$
$$= 1 + 1*2^4 + 1*2^1 + 1*2^3$$

$$= 1 + 16 + 2 + 8$$
$$= 27 \text{ rules}$$

Since $C = 1$ i.e. a unit factor, then the eq.4 can be generally stated (alternatively) as

$$R = 1 + \sum_{x=1}^{k} T^N_x \qquad eq.5$$

**Example 2:**

In this example with Figure 2(b), the ontology as shown is a little more complex than that of Figure 2(a). We shall begin the computational process by splitting the relational components into sub-units for analysis. So in a *bottom-up* approach, between Level L2 and L4, there are 5 parent classes of $C:C_1$ to $C_5$ with 3 at L2; and 2 at L3. From

$$R = 1 + \sum_{x=1}^{k} C_x T^N_x$$

we have
$$R = 1 + C_1 T^N_1 + [C_2 T^N_2 + C_3 T^N_3] + [C_4 T^N_4 + C_5 T^N_5]$$

would fit to cover all the parent class nodes $C$ at Level L2 and L3. Note that the square brackets is only to denote that the terms within it belongs to one parent class $C$ at Level L2.
Thus, for the term at $C_1$, the sub-unit shown in Fig. 2(bi) is:

$$C_1 T^N_1 = 1*2^4$$

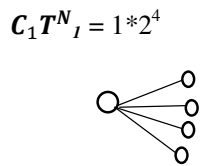

Fig. 2(bi)

At $C_2$, there are two terms based on the 2 sub-units to the horizontal as analysed in the Figure 2(b(ii) & b(iii)) below. Therefore

$$[C_2 T^N_2 + C_3 T^N_3] = 1*2^1 + 1*2^3$$

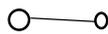         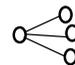

Fig. 2(b(ii))                Fig. 2(b(iii))

At $C_3$, there are also two terms to be derived from the 2 subclass units to the horizontal as shown in Figure2(b(iii) & b(iv)). That is

$$[C_4 T^N{}_4 + C_5 T^N{}_5] = 1*2^3 + 1*2^2$$

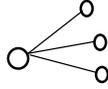
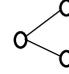

Fig. 2(b(iii))                    Fig. 2(b(iii))

In total summation,

$$R = 1 + 1*2^4 + 1*2^1 + 1*2^3 + 1*2^3 + 1*2^2$$

$$R = 1 + 16 + 2 + 8 + 8 + 4$$

$$R = 39 \text{ classified rules}$$

Systematically, the polynomial equation equally scaled to predict the accurate number of classified rules **R** even in *regular ontologies* e.g. Figure 1; see the example 3 below.

**Example 3a:**

In the Figure 1 above, there three parent class C at Level L2 that has a regular number of leaf-nodes *N = 2* at Level L3. Thus applying the polynomial equation

$$\mathbf{R = 1 + C_1 T^N{}_1 + C_2 T^N{}_2 + C_3 T^N{}_3}$$

We have,

$$R = 1 + 1*2^2 + 1*2^2 + 1*2^2$$

$$R = 1 + 4 + 4 + 4$$

$$R = 13 \text{ classified rules;}$$

**Example 3b:**

And the result of *R = 13* conforms to the prediction of the *Initialisation* algorithm

$$\mathbf{R = CT^N + 1}$$

With *C = 3, N = 2;* then

$$R = 3*2^2 + 1$$

$$R = 3*4 + 1$$

$$R = 13 \text{ classified rules}$$

The implication is that while in a regular ontology, the *condition-action* induction learning clauses for classification learning and recommendation are of equal number across the tree with respect to the regular number of nodes; in anon-regular ontology, the learning clauses arearbitrarily random according to the irregular structure of the ontology tree and each term in the *polynomial*. Based on the message received from the *sending* agent to a *receiving* agent (i.e. the *classifier* agent, see [16], [17], the *classifier* agent can then classify the environment (students) into one of the selective category according to number of nodes the students are pre-assessed on.

## 4. INDUCTION RULE KNOWLEDGE REPRESENTATION IN AGENTS

Figure 3 is a real model of a non-regular ontology in the domain of SQL learning structure in a teaching-learning environment.

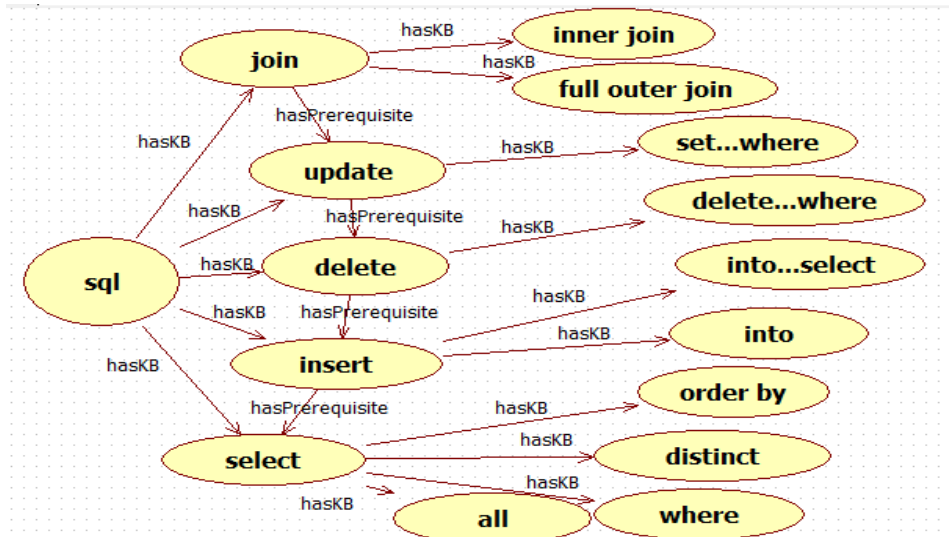

Fig. 3: A non-regular ontology in the domain of SQL.

With parent class or prerequisite C = 5 with

$C_1$ in a relation of leaf-nodes N = 4
$C_2$ in a relation of leaf-nodes N = 2
$C_3$ in a relation of leaf-nodes N = 1
$C_4$ in a relation of leaf-nodes N = 1
$C_5$ in a relation of leaf-nodes N = 2

The total number of classified rules as computed from

$$R = 1 + \sum_{x=1}^{5} C_x T^N{}_x$$

is therefore

$$R = 1 + C_1 T^N{}_1 + C_2 T^N{}_2 + C_3 T^N{}_3 + C_4 T^N{}_4 + C_5 T^N{}_5$$

$$R = 1 + 1*2^4 + 1*2^2 + 1*2^1 + 1*2^1 + 1.2^2$$

$$R = 1 + 16 + 4 + 2 + 2 + 4$$

$$R = 29 \text{ classified rules}$$

### 4.1 Combinations

The space of Bayesian networks is a combinatorial space [11]. Thus, let *c ϵ C*, *n ϵ N*, and *r ϵ R*; based on each term in the polynomial and *T = 2*, Table I thuspresents the combinatorial representation of the rules structure of the form $C_xN_yR_z$ of the *Examples 1*, *2*, *3a & b;* and *Figure 3. Example 3a(non-regular ontology)* is the equivalent of *Example 3b (a regular ontology)*.

Table I: Combinatorial Structure of Predicted Classified Rules

|  | **Example 1** | **Example 2** | **Example 3a** | **Example 3b** | **Figure 3** |
|---|---|---|---|---|---|
| 1. | c1n4r16 | c1n4r16 | c1n2r4 | C3N2R13 | c1n4r16 |
| 2. | c1n1r2 | c1n1r2 | c1n2r4 |  | c1n2r4 |
| 3. | c1n3r8 + 1 | c1n3r8 | c1n2r4 + 1 |  | c1n1r2 |
| 4. |  | c1n3r8 |  |  | c1n1r2 |
| 5. |  | c1n2r4 + 1 |  |  | c1n2r4 + 1 |
| **TOTAL** | 27 | 39 | 13 | 13 | 29 |

In the Table the all possible combinations for the classification of student users based on the boolean state classification parameter is given for the Figure 3 ontology.

Table II: Combinations and classification

| **Parent Class Node** | **Boolean Parameter Combinations** |
|---|---|
| @C1 | PPPP, PPPF, PPFF, PFFF, FFFF, FPPP, FFPP, FFFP, PFFP, PFPP, PPFP, FPPF, FPFF, FFPF, PFPF, FPFP |
| @C2 | PP, FF, PF, FP |
| @C3 | P, F |
| @C4 | P, F |
| @C5 | PP, FF, PF, FP |

In this system of multiagent, there is a classifier agent. Using the supervised learning approach, in Figures 4 and 5, we present the *input-output* pair of the predicate knowledge representation for agent classification learning: The *inputs* are the literal tokens passed to the classifier, and the *outputs* are the literals communicated by the classifier to the receiving agent. The Figures 4 and 5 are only two of the 16 cases (rules) for the first parent class node (the SELECT) out of the total rules R = 29 in the ontology tree (Fig. 3), as shown in Table II. With the graph in Figure 6, we then show the behaviour of the polynomial of the *non-regular ontology* tree rule prediction.

One of the conceptual meaning of learning is having to classify unknown data or feature into one of true or false class. In Figure 7, using predicate description logic, we show in Jason AgentSpeak language how agents classified students based on the probability notation P(Y | X): where X is the set of observations and Y the set of variables for prediction or diagnosis [11].

| Input | Output (URL) |
|---|---|
| pass(select_all) pass(select_where) pass(select_distinct) pass(order-by) | insert |

Fig. 4: Input-output pair representation for all positive (passed) examples.

| Input | Output (URL) |
|---|---|
| pass(select_all) fail(select_where) fail(select_distinct) fail(order-by) | select_where & select_distinct & order-by |

Fig. 5: Input-output pair representation for a mixed of both positive (passed) and negative (failed) examples.

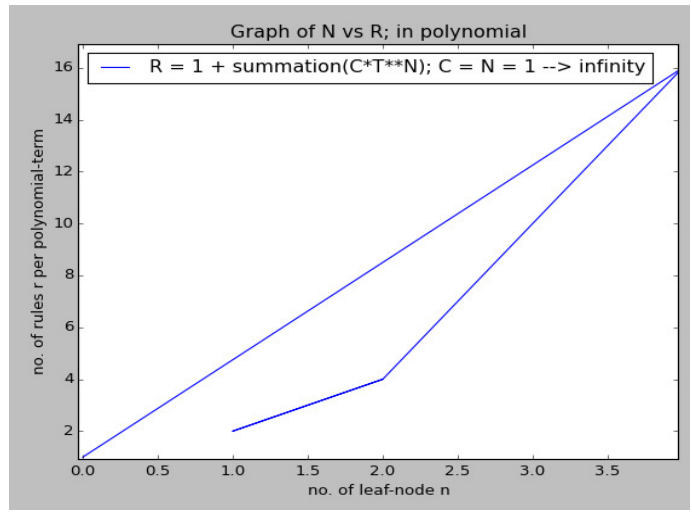

Fig. 6: Behaviour of the polynomial equation in the prediction of rules for a non-regular ontology structure as the leaf-nodes *N* are unordered.

The concepts with the *fail* predicate represents the diagnosis or misconception in students' cognitive knowledge. This also mandated the classification for *fail* predicate concepts in the "output" recommendations that is meant to *correct* the misconceptions (Fig. 5). The non-failure (*pass*) in Figure 4 is a *pass* classification for the next higher level concept in the tree. In Jason AgentSpeak [6], the following exemplar code snippet depicts the process of communication of these classification parameters to the classifier, learning and further communication by the classifier to the receiving agent for output of classified URL links.

```
@c1_ruleNon_Regular4
+!value(V)[source(agSupport)] : value("INSERT")
     &passed(... SELECT_ALL . . .)
     & failed (... SELECT_WHERE . . .)
     & failed (... SELECT_DISTINCT . . .)
     &failed(... ORDER_BY . . .)
     <- .send(agMaterial, achieve, hasKB(select, select_where));
     .send(agMaterial, achieve, hasKB(select, select_distinct));
     .send(agMaterial, achieve, hasKB(select, order_by)).
```

Fig. 7: Predicate logic formalism syntax structure for classification learning

## 5. CONCLUSION AND FURTHER WORK

This paper has demonstrated the estimation and prediction of the number of classification rules on a *non-regular* ontology structure using a system of polynomial equation, and subsequently knowledge representation of the predicted estimation for classification learning. While the polynomial equation can predict decision rules estimate for both a *regular* and *non-regular* ontologies, the *Initialisation* algorithm only predicts decision rules for regular ontologies. Classification learning is the process of specifying a pair of {X, Y} training examples for a system to learn in order for it to make selection of unknown input based on the given examples. In a learning environment of students, this process has been demonstrated with a non-regular ontology structure. The inputs for agents learning are the X predicate logic statements which are specified in the precinct of its pre-conditions for selective categorisation of some output Y specified in the body of the agent's plans. But it could be cumbersome to specify the rules for a big number of leaf-node N. Since there are systems of equations---the *Initialisation* algorithm and the Polynomial equation---the next stage in this work is to employ the use of agents for the production of their own classification rules based on the proven predictive equations.

**Authors**

Kennedy Efosa Ehimwenma is a PhD student in multiagent systems and semantic ontology. His research interests include multiagent cooperation, knowledge representation, predictive modelling, decision systems and agent classification learning. He is a member of the IEEE and British Computer Society.

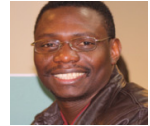

Assoc Professor Paul Crowther is the Deputy Head of the Department of Computing, Sheffield Hallam University. His expertise is in database systems with research interest in knowledge base systems, agents, and mobile learning. He is a Fellow of the Brtitish Computer Society.

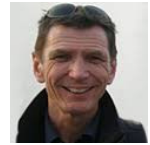

Dr Martin Beer is a Principle Lecturer in the Department of computing Sheffield Hallam University. His research interest and expertise are in multiagent systems, semantic web technology, database systems, and mobile learning. He is a Fellow of the Brtitish Computer Society.

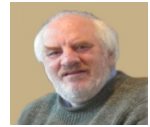

.